\documentclass[preprint,journal]{vgtc}       

\ifpdf
  \pdfoutput=1\relax                   
  \usepackage{graphicx}                
  \DeclareGraphicsExtensions{.pdf,.png,.jpg,.jpeg} 
\else
  \usepackage{graphicx}                
  \DeclareGraphicsExtensions{.eps}     
\fi%

\graphicspath{{figures/}{pictures/}{images/}{./}} 

\usepackage{microtype}                 
\PassOptionsToPackage{warn}{textcomp}  
\usepackage{textcomp}                  
\usepackage{mathptmx}                  
\usepackage{times}                     
\usepackage{cite}                      
\usepackage{tabu}                      
\usepackage{booktabs}                  
\usepackage{amsmath}
\usepackage[ruled,vlined]{algorithm2e}

\usepackage{subfig}

\usepackage{xcolor} 

\DeclareMathAlphabet{\mathcal}{OMS}{cmsy}{m}{n}

\DeclareMathOperator*{\argmin}{arg\,min}

\onlineid{1033}

\vgtccategory{Research}
\vgtcpapertype{algorithm/technique}

\title{Redirected Walking in Static and Dynamic Scenes\\Using Visibility Polygons}

\author{Niall L. Williams, \textit{Student Member, IEEE}, Aniket Bera, \textit{Member, IEEE}, and Dinesh Manocha, \textit{Fellow, IEEE}}
\authorfooter{
\item
 Niall L. Williams, Aniket Bera, and Dinesh Manocha are with the University of Maryland, College Park. E-mail: \{niallw$\vert$ab$\vert$dm\}@cs.umd.edu.
}

\shortauthortitle{Biv \MakeLowercase{\textit{et al.}}: Global Illumination for Fun and Profit}

\abstract{
We present a new approach for redirected walking in static and dynamic scenes that uses techniques from robot motion planning to compute the redirection gains that steer the user on collision-free paths in the physical space. 
Our first contribution is a mathematical framework for redirected walking using concepts from motion planning and configuration spaces.
This framework highlights various geometric and perceptual constraints that tend to make collision-free redirected walking difficult. 
We use our framework to propose an efficient solution to the redirection problem that uses the notion of visibility polygons to compute the free spaces in the physical environment and the virtual environment. 
The visibility polygon provides a concise representation of the entire space that is visible, and therefore walkable, to the user from their position within an environment.
Using this representation of walkable space, we apply redirected walking to steer the user to regions of the visibility polygon in the physical environment that closely match the region that the user occupies in the visibility polygon in the virtual environment.
We show that our algorithm is able to steer the user along paths that result in significantly fewer resets than existing state-of-the-art algorithms in both static and dynamic scenes.
Our project website is available at \href{https://gamma.umd.edu/vis_poly/}{\textcolor{blue}{\texttt{https://gamma.umd.edu/vis\_poly/}}}.
} 


\CCScatlist{ 
 \CCScat{K.6.1}{Management of Computing and Information Systems}%
{Project and People Management}{Life Cycle};
 \CCScat{K.7.m}{The Computing Profession}{Miscellaneous}{Ethics}
}

\teaser{
  \centering
  \includegraphics[width=\linewidth]{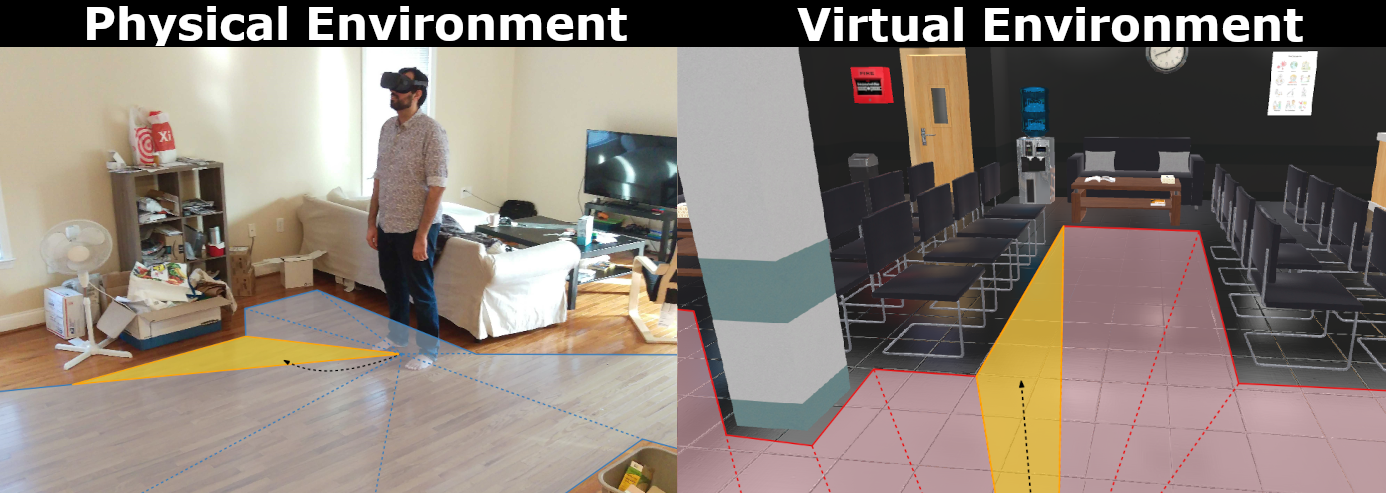}
  \caption{A visualization of the geometric reasoning that our redirection controller performs on every frame in order to steer the user in the physical space. First, the controller computes the visibility polygon for the user's physical and virtual locations (the regions bounded by the blue and red edges, respectively). Next, the controller computes the region of space (part of the red visibility polygon) in front of the user that the user is walking towards in the virtual environment (yellow region in the right image). By comparing the areas of the regions, our controller computes the region in the physical space (yellow region in the left image) that is most similar to the virtual region the user is heading towards. Finally, the controller applies redirected walking gains to steer the user to walk towards the highlighted region in the physical space. Black dashed arrows indicate the user's trajectory in the environment. Our algorithm yields significantly fewer resets with physical obstacles than prior algorithms.}
	\label{fig:teaser}
}

\vgtcinsertpkg

\begin{document}



\maketitle

\label{sec:introduction}
\section{Introduction}

Natural walking as a means to explore virtual environments (VEs) is generally preferred over other artificial locomotion interfaces such as flying \cite{usoh1999walking} or teleportation since natural walking improves the user's sense of presence and task performance in the VE.
Redirected walking (RDW)~\cite{razzaque2005redirected} is a locomotion interface that affords natural walking in virtual reality by imperceptibly steering the user along physical paths that differ from their virtual counterparts.
Over the years, many researchers have developed different algorithms (known as \textit{RDW controllers}) that apply RDW to steer the user to avoid collisions with physical obstacles that they cannot see or detect.
If a collision is imminent, the RDW system applies a ``reset'' to reorient the user away from the nearby obstacle.
Although significant progress has been made, even the best controllers are unable to guarantee a reset-free experience in arbitrary static and dynamic environments.

Redirection controllers generally fall into one of three categories: reactive, predictive, or scripted \cite{nilsson201815}.
Reactive controllers steer the user based only on information available at the current or previous frames.
Predictive controllers steer the user based on the user's predicted future movement in the VE, and scripted controllers steer the user as they walk along a pre-defined path in the VE.
Reactive controllers are usually preferred since they can be deployed in arbitrary environments and often require little to no additional setup (such as environment pre-processing).
However, their ability to steer the user away from obstacles is usually lower than that of predictive or scripted controllers since reactive controllers must operate using less information.
Predictive controllers can outperform reactive controllers, but they rely on accurate predictions of the user's movements.
Recently, a new paradigm of redirection controllers based on the concept of alignment has emerged.
Instead of steering the user away from physical obstacles, alignment-based controllers steer the user in an attempt to minimize the difference between the physical and virtual environments.
These controllers have been shown to be effective for lowering the number of resets \cite{williams2021arc} and enabling passive haptics for increased immersion \cite{thomas2020reactive,thomas2020towards}.

Much of the prior work on RDW controllers focuses on users exploring static physical and virtual environments.
Some work that instead focuses on dynamic scenes mostly looks at multi-user RDW systems \cite{bachmann2019multi,dong2020dynamic,azmandian2017evaluation}.
Chen et al. \cite{chen2018redirected} proposed ideas on how redirection controllers can steer the user away from moving physical obstacles.
Though there has been work on dynamic physical scenes as well as multi-user systems, there are no known solutions that work well in all dynamic environments with no assumptions on obstacles' motions.

\textbf{Main Results:} 
We provide a mathematical framework for the redirected walking problem, as well as a novel approach to redirected walking in static and dynamic environments.
We frame the redirection problem as optimizing the user's path in the physical environment by applying the appropriate redirection at each frame.
Our formulation makes it easy to develop effective controllers and conduct rigorous analyses to better understand a redirection controller's behavior.
Using this formulation of the redirection problem, we also introduce a RDW controller that uses a novel alignment metric based on visibility polygons.
That is, given the layouts of the physical and virtual environments and the user's location in them, our controller computes the region of walkable space visible from the user's position in both environments.
With these representations of the local space that is available to the user, our controller steers the user towards a region in the physical space that is most similar to the region they are approaching in the virtual space.
We found that our controller is able to significantly reduce the number of times that the user has to reset their orientation when they come too close to physical obstacles.
Our main contributions in this work include:
\begin{itemize}
\setlength\itemsep{.25em}
    \item A mathematical framework for the redirected walking problem based on concepts from robot motion planning. We formulate it as an optimization problem, where the aim is to transform a virtual path to an optimal physical path, subject to constraints depending on the user and the information available.
    \item A novel redirected walking controller that steers the user based on alignment, with heuristics derived from visibility polygons. Our algorithm achieves significantly fewer resets than the current state-of-the-art controllers.
    \item Comparisons of our algorithm to the state-of-the-art algorithms in both static and dynamic simulated, single-user scenes.
\end{itemize}

\label{sec:background}
\section{Prior Work and Background}
Locomotion is a fundamental problem in virtual reality (VR) since the user wishes to explore a VE that is usually much larger than the physical environment (PE) they are in.
Furthermore, the locations of obstacles in the PE are usually not the same as the locations of obstacles in the VE.
This creates a problem when the user wishes to travel along a collision-free virtual path but doing so may force them to walk into unseen physical obstacles.
Many different techniques for locomotion have been developed \cite{sun2018towards,steinicke2013human,interrante2007seven,usoh1999walking,peck2011evaluation,razzaque2005redirected}, but in this work we focus on the RDW technique developed by Razzaque et al. \cite{razzaque2001redirected}.
A recent review of VR locomotion interfaces can be found in \cite{luca2021locomotion}.

By slowly rotating or translating the VE around the user while they walk, RDW allows users to explore virtual environments while located in smaller physical environments \cite{razzaque2005redirected}.
These rotations and translations are controlled by parameters called \textit{gains}.
In order to remain on their intended virtual path, users will subconsciously adjust their physical path to counteract the VE movements. 
The three main gains are translation, rotation, and curvature gains.
Translation gains translate the VE around the user while they walk, causing their physical path to be longer or shorter than their virtual path, depending on the direction of the VE translation.
Rotation gains rotate the VE around the user while they turn in place, which causes their physical rotation to be larger or smaller than their virtual rotation one, depending on the direction that the VE rotates.
Similarly, curvature gains rotate the VE around the user while they are walking, which causes them to veer on a physical path with a different curvature than their virtual one.
The faster the VE rotates or translates, the more the user's physical path will deviate from their virtual path, and the easier it will be to steer the user away from physical obstacles.
However, it is important that the gains are not large enough that the user can perceive the rotations since this will make it harder to explore and can lead to simulator sickness \cite{steinicke2009estimation}.
Other gains such as bending gains \cite{langbehn2017bending} have been developed, but they are less well-understood than translation, curvature, and rotation gains so we do not consider them in this work.

\subsection{Redirection Controllers}
\label{subsec:controllers}
A redirection controller is an algorithm that applies RDW gains to steer a user along a physical path, with the intention of minimizing the number of times the user collides with a physical obstacle \cite{nilsson201815}.
A controller consists of a \textit{steering} component and a \textit{resetting} component.
As the name suggests, the steering component is responsible for applying gains to steer the user in the physical space while walking.
The resetting component is responsible for initiating a ``reset,'' wherein the user's virtual movements are disabled until they turn in place to reorient themself in the physical world.
Resets are initiated when the user gets too close to any physical obstacle.

Many redirection controllers operate on different assumptions and with different amounts of information available to them.
Consequently, controllers have traditionally been classified as reactive, predictive, or scripted \cite{nilsson201815}.
Reactive controllers steer the user according to the information available at the current frame or any prior frames.
These controllers are typically designed to function in arbitrary environments, with little to no pre-processing or setup required.
Examples of reactive controllers include steer to center \cite{razzaque2005redirected}, steer to orbit \cite{razzaque2005redirected, hodgson2013comparing}, steer to multiple targets \cite{razzaque2005redirected}, and controllers based on reinforcement learning \cite{strauss2020steering,chang2019redirection,lee2019real} or artificial potential fields \cite{thomas2019general,messinger2019effects,bachmann2019multi}.
Predictive controllers steer the user according to the information available on the current/prior frames and a prediction of the user's future path in the VE.
These types of controllers can perform better than reactive controllers, but their performance depends on the accuracy of the path prediction.
Examples of predictive controllers include those developed by Zmuda et al. \cite{zmuda2013optimizing}, Nescher et al. \cite{nescher2014planning}, and Dong et al. \cite{dong2020dynamic}.
Scripted controllers are controllers that steer the user as they travel along designated paths in the VE \cite{azmandian2018design, yu2018experiencing}.
Scripted controllers usually result in the fewest resets, but they require the researchers to design and plan the environments carefully.
As a result, these controllers cannot be generalized to arbitrary environments.

Although the RDW community has developed this taxonomy for redirection controllers, more recent controllers that involve more complex algorithms or use new types of information are not easily categorized as reactive, predictive, or scripted.
Thomas et al. \cite{thomas2020reactive,thomas2020towards} and Williams et al. \cite{williams2021arc} introduced redirection controllers that leverage the concept of alignment.
\textit{Alignment} is the concept of comparing the physical and virtual environments according to some environment feature(s) (such as the location of a user relative to an object).
In other words, alignment measures the similarity of the two environments according to these environment features.
It should be noted that other researchers have used ideas similar to alignment by editing the VE to match the physical one \cite{simeone2020space} or by making assumptions about the user's motion based on the VE \cite{zmuda2013optimizing}.
The alignment-based controllers developed by Thomas et al. and Williams et al. steer the user with RDW gains, but the steering decisions are guided by the degree of similarity (alignment) of the user in the physical and virtual environments.
Furthermore, some controllers make implicit assumptions about the virtual reality system and incorporate these assumptions into the steering policy.
Williams et al. \cite{williams2021arc} assume that the user walks along a collision-free path in the VE, while controllers based on reinforcement learning train an algorithm that implicitly learns a model of the user's locomotion behavior and steers the user according to this model.

\subsection{Motion Planning and Visibility Polygons}

In the field of robotics, motion planning is the problem of moving a robot from an initial state to a goal state through a series of valid configurations that avoid collisions with obstacles \cite{lavalle2006planning}.
For a robot with \textit{n} degrees of freedom, its configuration space (denoted $\mathcal{C}$) is an $n$-dimensional manifold, where each point in the manifold corresponds to a configuration of the robot.
The configuration space describes the set of all states that a robot can be in.
In order to successfully navigate from a starting position to a goal position, the robot must find a set of configurations that takes it from the starting position to the goal position.
This can be formulated as finding a continuous path of valid configurations through $\mathcal{C}$.
Some common desirable traits of such a path are that it yields the shortest path and that the robot trajectory is smooth, without many oscillations as it travels along this path.

There is considerable work on developing motion planning algorithms for static and dynamic environments. 
Search-based planners discretize the state space (the set of all possible states) and employ search algorithms to find a path from the start to the goal.
An example of a search-based motion planning algorithm is the A* algorithm \cite{hart1968formal}.
Sampling-based planners operate by randomly sampling the configuration space in order to build a valid path.
Such algorithms can usually quickly find valid solutions, but their solutions are usually not the most efficient \cite{elbanhawi2014sampling}.
Potential field methods uses attractive and repulsive forces to guide the robot through the environment \cite{khatib1986real}.
These planners are easy to implement but are susceptible to getting the robot trapped in local minima of the potential function.
Planning algorithms may also use geometric representations, such as visibility graphs and cell decomposition, to reason about the environment, detect collisions, and compute collision-free paths \cite{de1997computational}.
Motion planning algorithms may also use optimization to to handle dynamic obstacles and compute smooth trajectories \cite{park2012itomp, ratliff2009chomp}.
Optimization-based approaches are advantageous in that they can more easily handle complex, high-dimensional state spaces.
Dynamic motion planning is the problem of computing a collision-free path in an environment with moving obstacles.
A popular approach to dynamic motion planning is the use of velocity obstacles to reason about collision-free paths in terms of velocity \cite{fiorini1998motion, van2011reciprocal}.

In our approach, we use techniques from motion planning literature to compute collision-free paths for the user in the physical space and the virtual space. 
In particular, we perform geometric reasoning in each of these 2D spaces by using \textit{visibility polygons}.
The visibility polygon for a point $p$ is the set of all points in the plane that are visible from $p$.
The visibility polygon thus encapsulates the entire region of space that has line-of-sight visibility from the point $p$.
Depending on the layout of the space, the visibility polygon may be unbounded.
If we consider a human (or robot) observer located at position $p$, the visibility polygon can be thought of as the entire region of space that the observer can see.
The visibility polygon can be used to compute the free space corresponding to a point robot in an environment with polygonal obstacles \cite{guibas1995robot}.
Thus, the visibility polygon provides a well-defined region that we can use to compute a local path in the environment to move the user closer to their goal without any collisions.

Since human locomotion is largely dominated by the information that is immediately available to the person \cite{montello2005navigation}, we use the visibility polygon to perform geometric reasoning in a user's local surroundings.
In the architectural design literature, researchers have used visibility polygons (which they refer to as ``isovists'') to describe an environment and then studied how people's locomotion patterns change as the environment structure changes \cite{benedikt1979take}.
Wiener et al. \cite{wiener2007isovist} showed that the complexity of the visibility polygon (characterized by its jaggedness) was correlated with an observer's task performance and locomotion speed in the environment.
Christenson et al. \cite{christenson2010registering} introduced occlusion maps, which are maps that define the regions visible over multiple points in an environment, to study how the layout of an environment changes as the observer moves to a new position.
Within the RDW community, Zank et al. \cite{zank2017optimized} used visibility polygons to predict the areas of the VE that the user might walk towards next.




\label{sec:methods}
\section{Redirected Walking Using Visibility Polygons}
To aid in the exposition of our redirection controller, we begin with formal definitions and a mathematical formulation of the RDW problem.
A formal characterization of the RDW problem provides us with a framework that we can use to conduct more rigorous analyses of and reasoning about RDW steering algorithms.
Note that the definitions we use are adapted from those used in the robot motion planning literature \cite{lavalle2006planning}.


\subsection{Definitions and Notation}
\label{subsubsec:definitions}
In virtual reality, the user is simultaneously located in a PE, $E_{phys}$, and a VE, $E_{virt}$.
For each environment, the user state describes their location $p$ and heading direction $\theta$ in the environment.
We denote the user's state as the pair $q = \{p, \theta\}$.
The user's state at a particular time $t$ is denoted by $q^t = \{p^t, \theta^t\}$.
Thus, the user's state at time $t$ in $E_{phys}$ is denoted by $q^t_{phys} = \{p^t_{phys}, \theta^t_{phys}\}$.
The path that a user walks along in an environment is represented by an ordered set of states $Q = \{q^0, q^1, ..., q^{t}\}$.
For brevity, we denote the physical and virtual paths as $path_{phys} = \{ q^0_{phys},..., q^t_{phys}\}$ and $path_{virt} = \{ q^0_{virt},..., q^t_{virt}\}$, respectively.

The user's state in an environment is also referred to as their configuration.
The configuration space $\mathcal{C}$ (or $\mathcal{C}$-space for short) is the set of all possible configurations of the user.
Some configurations in the $\mathcal{C}$-space correspond to the user colliding with an obstacle in the environment.
The obstacles in an environment occupy the obstacle region, $\mathcal{O} \subset E$.
The set of all colliding configurations $q$ is the obstacle space $\mathcal{C}_{obs}$:
\begin{equation} \label{eqn:c_obs}
    \mathcal{C}_{obs} = \{q\in\mathcal{C} \ | \ q \cap \mathcal{O} \neq \emptyset\}.
\end{equation}
That is, the obstacle space $\mathcal{C}_{obs}$ is the set of all configurations $q$ for which the user intersects with an obstacle in the obstacle region $\mathcal{O}$.
The free space $\mathcal{C}_{free}$ is all of the other configurations that are not in the obstacle region.
We define the free space as $\mathcal{C}_{free} = \mathcal{C} \setminus \mathcal{C}_{obs}$.
The free space and obstacle space of the PE are denoted $Free_{phys}$ and $Obs_{phys}$, while the free and obstacle spaces of the VE are denoted $Free_{virt}$ and $Obs_{virt}$.
Finally, note that $E_{phys} = Free_{phys} \cup Obs_{phys}$ and $E_{virt} = Free_{virt} \cup Obs_{virt}$.

\subsection{Redirected Walking and Configuration Spaces}
\label{subsubsec:problem_statement}
\begin{figure}[t]
    \centering
    \includegraphics[width=.47\textwidth]{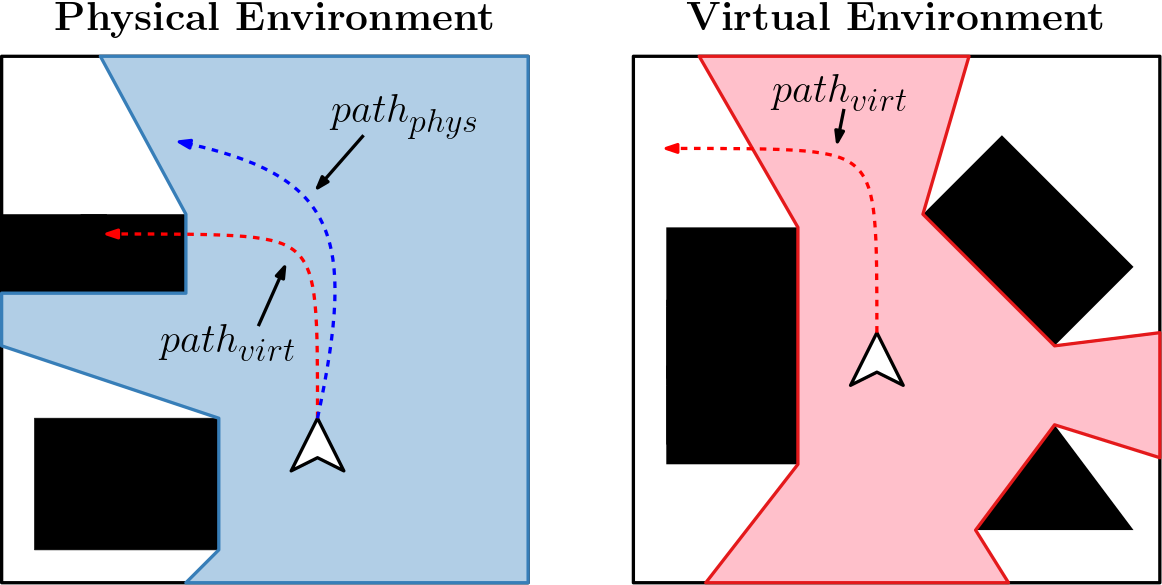}
    \caption{Visualization of the redirected walking problem. If the user tries to walk on the virtual path $path_{virt}$ with no redirection applied, they will collide with the obstacle to their left in the physical space. After applying redirection, the user instead walks along $path_{phys}$ and avoids any collisions. The free spaces $Free_{phys}$ and $Free_{virt}$ are shown in blue and red, respectively.}
    \label{fig:rdw_mental_model}
\end{figure}

Redirected walking is typically implemented by rotating the VE around the user as they walk, so it is natural to think about a redirection controller as an algorithm that rotates the VE according to some criteria on every frame.
Instead, our goal is to frame redirection controllers in terms of the simultaneous computation of collision-free trajectories in the physical and virtual spaces.
We can visualize this as superimposing the user's virtual path onto their physical location and applying RDW gains that transform the superimposed path such that the user avoids any obstacles that the path intersects with (see \autoref{fig:rdw_mental_model}).
In our figures, black shapes represent obstacles, white space is any walkable region in the environment, and the colored regions represent the free space (\textit{i.e.} the subset of the walkable region visible from the user's position).

With our goal in mind, we can now formally describe the redirection problem using the definitions from \autoref{subsubsec:definitions}.
Given two environments $E_{phys}$ and $E_{virt}$, the user's configuration in both environments are $q_{phys}$ and $q_{virt}$, respectively.
The user explores $E_{virt}$ by following a collision-free path $path_{virt} \in Free_{virt}$.
This path corresponds to some physical path $path_{phys} \in E_{phys}$, for which it is possible that $path_{phys}~\cap~Obs_{phys}~\neq~\emptyset$.
If ${path_{phys} \cap Obs_{phys} \neq \emptyset}$, we wish to find some new optimal path $path^*_{phys}$ such that $path^*_{phys} \cap Obs_{phys} = \emptyset$, i.e. $path^*_{phys} \in Free_{phys}$.
Let \texttt{RDW($path$)} be a generic redirection function which applies translation ($g_t$), curvature ($g_c$), or rotation ($g_r$) gains at each timestep $q^t \in path$ to yield a new path $path^*$.
The redirection problem is thus to find some best function \texttt{RDW*()} such that $\texttt{RDW*(}path_{virt}\texttt{)} = path^*_{phys}$.
We seek to develop a redirection controller that executes \texttt{RDW*()}.
\texttt{RDW()} is subject to multiple constraints:
\begin{enumerate}
    \item \textbf{Maximum redirection constraint:} Since RDW is limited by human perception, the gains applied by \texttt{RDW()} must be bounded by the user's empirically measured perceptual thresholds. This limits how much redirection can be done at any given moment and makes it more difficult to avoid $Obs_{phys}$. 
    
    \item \textbf{Geometric deviation constraint:} \texttt{RDW($path_{virt}$)} results in a path $path_{phys}$ that is geometrically different from $path_{virt}$, where translation gains change the length and rotation and curvature gains change the curvature of $path_{phys}$ relative to $path_{virt}$. Stronger gains correspond to a greater deviation between $path_{phys}$ and $path_{virt}$. Stronger redirection gains have a higher chance of inducing simulator sickness in the user, so we wish to compute $path_{phys} \in Free_{phys}$ with the weakest gains possible. Geometrically, this is defined as the path with the lowest deviation from $path_{virt}$.
    
    \item \textbf{Information constraint:} Oftentimes, the user's future virtual path $path_{virt}$ is not known, which makes optimizing $path_{phys}$ difficult. Additionally, the user's PE may not be completely known, depending on the capabilities of the virtual reality system.
\end{enumerate}
Due to the above constraints, developing the perfect redirection function \texttt{RDW*()} for all $path_{virt}$ is very difficult.

Since it is difficult to develop \texttt{RDW*()}, we must rely on our \texttt{RDW()} function in conjunction with a resetting function \texttt{reset()}.
The resetting function is responsible for prompting the user to stop walking and reorienting them to a safe configuration in $E_{phys}$.
Similarly to \texttt{RDW()}, \texttt{reset()} applies redirection gains to transform $path_{virt}$ such that the resulting physical path is valid, i.e. ${path_{phys} \in Free_{phys}}$.
However, \texttt{reset()} differs from \texttt{RDW()} in that \texttt{reset()} alters $path_{virt}$ in order to yield a safe $path_{phys}$.
In practice, this is performed by having the user turn $360^\circ$ in place in $E_{virt}$, while reorienting by some smaller angle in $E_{phys}$ due to rotation gains.
Thus, \texttt{reset()} inserts configurations into $path_{virt}$ that cause the user to turn in place in $E_{virt}$ while applying rotation gains $g_r$ to alter the user's corresponding physical rotation.


\subsection{Finding \texttt{RDW*()} Using Visibility Polygons}
In this section, we present a new redirection controller that uses the notion of visibility polygons to compute \texttt{RDW()}.
Our goal is to steer the user along a path in $Free_{phys}$.
The visibility polygon computed in $E_{phys}$ at the user's position $p_{phys}$ is a representation of $Free_{phys}$ that can be computed in $O(n \ logn)$ time \cite{de1997computational,suri1986worst}, where $n$ is the number of segments that define the boundaries of obstacles in the environment.
Thus, in order to take advantage of motion planning techniques, our controller is based on the visibility polygon.

\begin{figure} 
    \centering
  \subfloat[Configurations in physical and virtual environments that lead to a collision. After the user walks forward (rightmost image), they are still in $Free_{virt}$ but they are no longer inside $Free_{phys}$, indicating that there was a collision with a physical obstacle.\label{1a}]{%
       \includegraphics[width=1\linewidth]{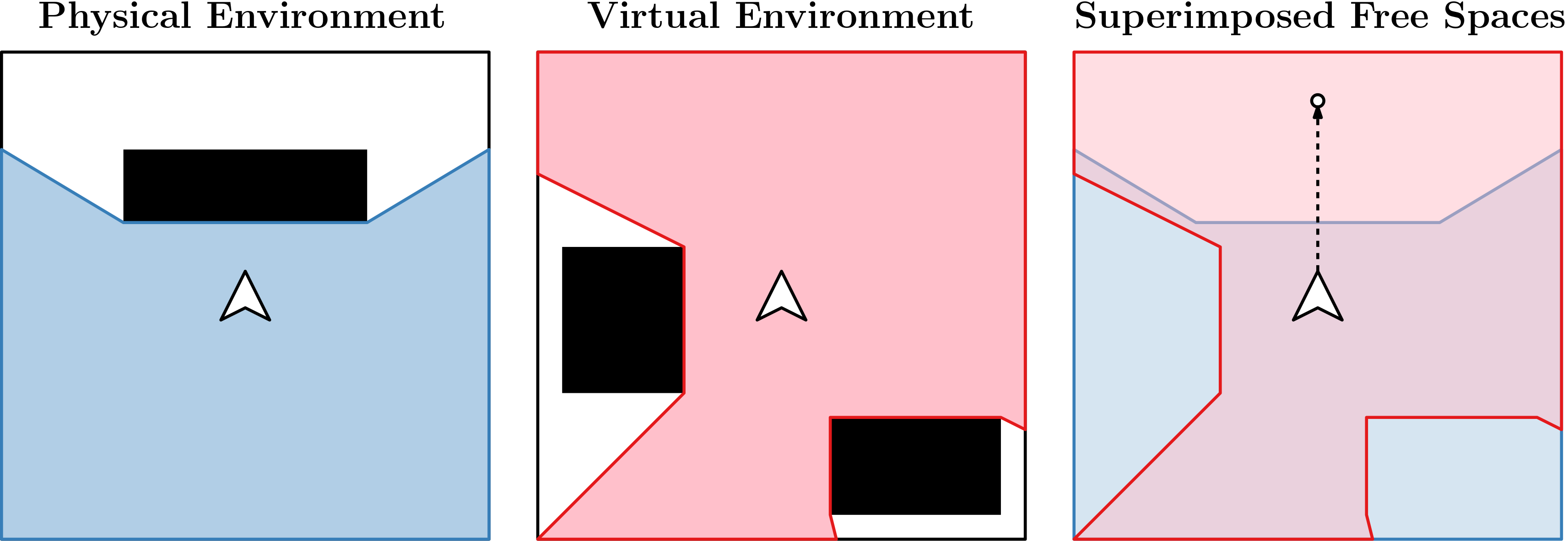}}
    \\
  \subfloat[Free space configurations in physical and virtual environments that do not lead to a collision. After the user walks forward (rightmost image), they are still within $Free_{virt}$ \textit{and} $Free_{phys}$, indicating that there was no collision along the physical path that the user travelled.\label{1b}]{%
        \includegraphics[width=1\linewidth]{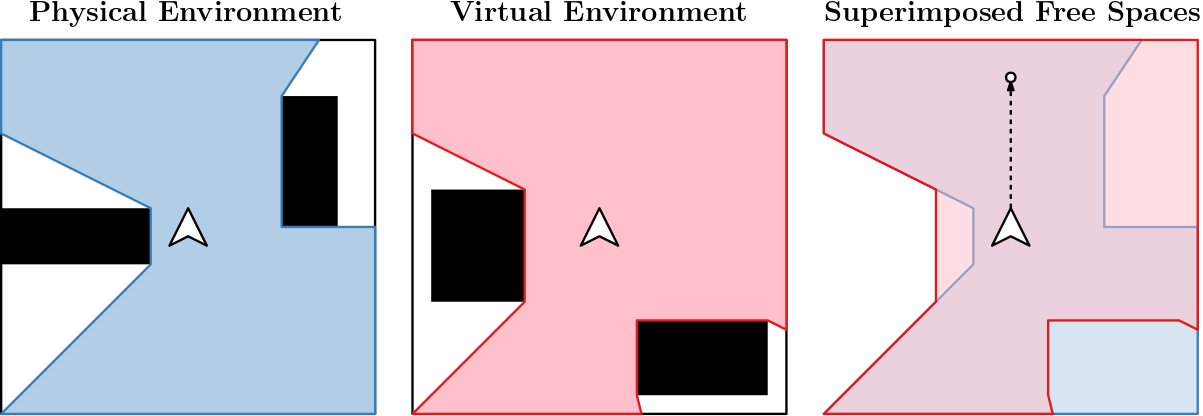}}
  \caption{A visualization of the superimposition of the two free spaces, $Free_{phys}$ (blue) and $Free_{virt}$ (red). Regions of $Free_{virt}$ that do not overlap with $Free_{phys}$ signify regions of the virtual environment that the user cannot walk to without colliding with a physical obstacle. Our controller aims to steer the user in $E_{phys}$ such that $Free_{phys}$ and $Free_{virt}$ overlap in the region that the user is walking towards.}
  \label{fig:superimpose} 
\end{figure}

Considering the problem statement in \autoref{subsubsec:problem_statement}, our controller's goal is to apply a function \texttt{RDW($path_{virt}$)}$=path_{phys}$ such that $path_{phys}$ is in the physical visibility polygon (which represents $Free_{phys}$).
Our redirection controller (which executes \texttt{RDW()}) is not predictive, so it does not have access to the user's future virtual path $path_{virt}$.
This makes it effectively impossible to compute the optimal physical path $path^*_{phys}$ as described in \autoref{subsubsec:problem_statement}, since we cannot directly transform $path_{virt}$ to make it lie within $Free_{phys}$.
To resolve this, we reframe the redirection problem slightly and use alignment in our redirection function \texttt{RDW()}.
Instead of superimposing $path_{virt}$ onto $E_{phys}$, we superimpose $Free_{virt}$ onto $Free_{phys}$, centered on the user (both $Free_{virt}$ and $Free_{phys}$ are represented by visibility polygons).
When the user walks in $E_{virt}$, their path in the superimposed $Free_{virt}$ will correspond to some path within $E_{phys}$, and hopefully within $Free_{phys}$.
Our controller's goal is now to apply redirection such that $Free_{phys}$ is transformed to match $Free_{virt}$ as closely as possible.
The intuition here is that the more similar $Free_{phys}$ is to $Free_{virt}$, the higher the chance that the user's next virtual configuration $q^{t+1}_{virt}$ will correspond to a valid physical configuration $q^{t+1}_{phys} \in Free_{phys}$.
This process of superimposing $Free_{phys}$ and $Free_{virt}$ is shown in \autoref{fig:superimpose}.


We guide this free space matching process using our alignment metric, which we define as the area of the free space in front of the user (see \autoref{subsubsec:get_most_similar_slice}).
Note that while we framed the problem as transforming $Free_{phys}$, in implementation, we simply apply RDW gains to change the user's physical configuration such that their new $Free_{phys}$ is more similar to $Free_{virt}$.
The pseudocode that our redirection controller executes on every frame is shown in \autoref{alg:rdw}.
We now provide details of each step of the algorithm.

\begin{figure*}[t]
    \centering
    \includegraphics[width=.99\textwidth]{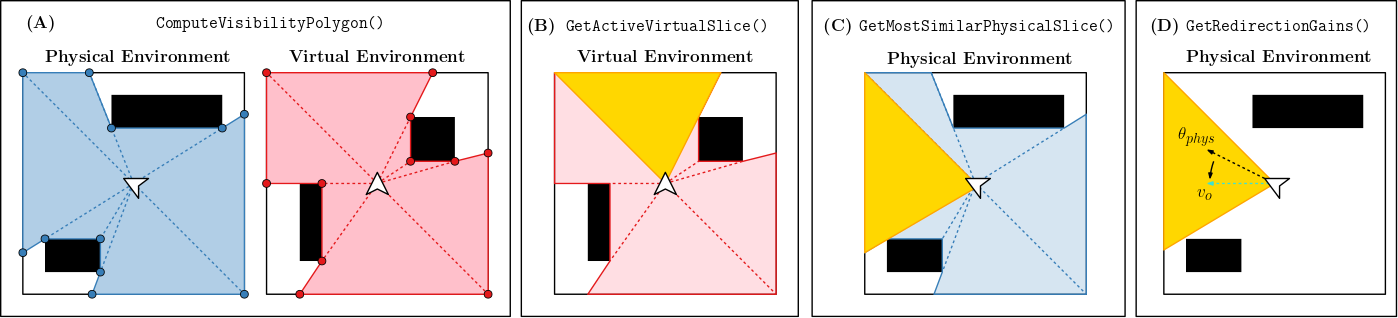}
    \caption{An overview of our redirection controller based on visibility polygons. 
    \textbf{(A)} We compute the visibility polygon corresponding to the user's position in both the physical (blue) and virtual (red) environments. After the visibility polygons are computed, they are divided into regions called ``slices'' which we use later in our approach to measure the similarity of the two polygons. 
    \textbf{(B)} The ``active slice'' in the virtual environment is computed. This is the slice of the virtual visibility polygon that the user is walking towards (shown in yellow). 
    \textbf{(C)} The corresponding slice in the physical environment that is most similar to the active slice is computed. 
    Similarity is measured using slice area.
    \textbf{(D)} Redirected walking gains are applied according to the user's heading to steer them in the direction of the most similar physical slice that was computed in step (C).}
    \label{fig:alg_overview}
\end{figure*}


\begin{algorithm}[ht]
\SetAlgoLined
\KwResult{Redirection gains $g_r, g_t, g_c$ to apply on the current frame.}
 $P_{phys}$ = \texttt{ComputeVisibilityPolygon($E_{phys}, p_{phys}, \theta_{phys}$)}\;
 $P_{virt}$ = \texttt{ComputeVisibilityPolygon($E_{virt}, p_{virt}, \theta_{virt}$)}\;
 $s^{virt}$ = \texttt{GetActiveVirtualSlice($P_{virt},\theta_{virt}$)}\;
 $s^{phys}$ = \texttt{GetMostSimilarPhysicalSlice($P_{phys},s^{virt}$)}\;
 $g_r,g_t,g_c$ = \texttt{GetRedirectionGains($s^{virt}, s^{phys}, \theta_{phys}$)}\;
 \caption{Redirection Gain Computation}
 \label{alg:rdw}
\end{algorithm}

\subsubsection{\texorpdfstring{\texttt{ComputeVisibilityPolygon($E, p, \theta$)}}{}}
Given an environment $E$ and a position $p$ and heading $\theta$ in that environment, we compute the visibility polygon using the algorithm described by Suri et al. \cite{suri1986worst} (note that $\theta$ is not used in this initial computation).
The visibility polygon $P$ is defined by a kernel $k$ and a set of vertices ${\{v_0, v_1, ..., v_{n-1}\}}$.
Here, $k$ is the position of the observer (i.e., the position $p$ in the environment $E$).
The set of vertices defines the edges of $P$, where consecutive vertices $v_i$ and $v_{i+1}$ form an edge.

\begin{figure}[!t] 
    \centering
  \subfloat[A visibility polygon, its vertices, and its slices $\{ s_1, s_2, ..., s_7 \}$.\label{fig:vis_poly_sliced}]{%
       \includegraphics[width=0.45\linewidth]{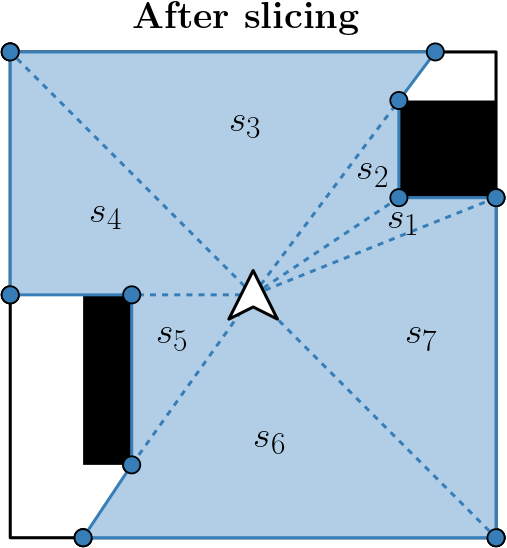}}
    \hfill
  \subfloat[A diagram showing some of the components that are computed for every slice $s$. Note that we do not show the average length $s_l$ or the kernel position \textit{s}{[0]} in order to reduce visual clutter.\label{fig:slice_anatomy}]{%
        \includegraphics[width=0.45\linewidth]{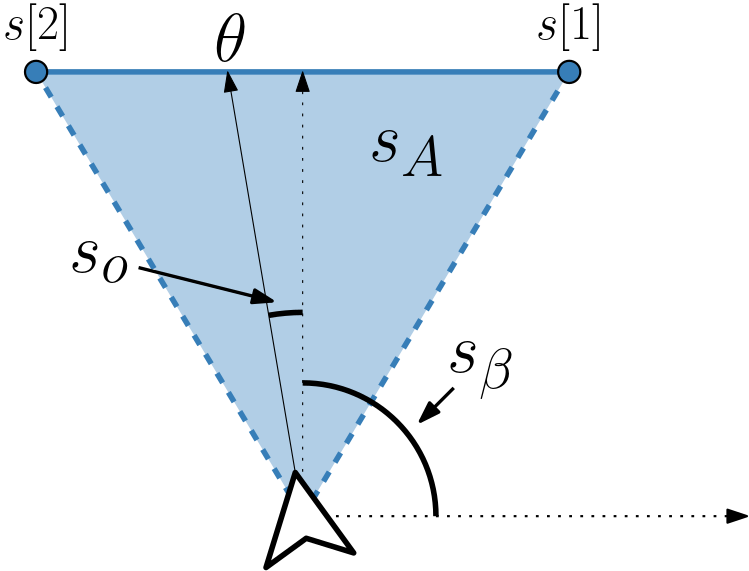}}
  \caption{A visibility polygon after its slices are computed (\autoref{fig:vis_poly_sliced}) and the composition of one slice (\autoref{fig:slice_anatomy}).}
  \label{fig:slice_stuff} 
\end{figure}

Once $P$ is computed, it is divided into ``slices'' $s_i$ around $k$ according to the order and position of the vertices of $P$.
The vertices are sorted in counterclockwise order around $k$, and slices are defined as the triangles formed by triplets of points $\{ k, v_i, v_{i+1} \}$.
If the points $\{ k, v_i, v_{i+1} \}$ are colinear, the slice is instead defined by the next vertex $v_{i+x} \in P$ such that $x>1$ and $\{ k, v_{i+1}, v_{i+x} \}$ are not colinear.
A diagram of these slices is shown in \autoref{fig:vis_poly_sliced}.

For each slice $s$, the following attributes are computed:
\begin{itemize}
    \item \textbf{Slice vertices:} The vertices $\{ s[0], s[1], s[2] \}$ that define the slice. Note that $s[0]=k$ and $\{ s[1], s[2] \} \subset P$.
    \item \textbf{Slice bisector ($s_\beta$):} The angle $\phi \in [0, 2\pi)$ that bisects angle $\angle s[1]s[0]s[2]$.
    \item \textbf{Average length ($s_l$):} The average length of the two line segments connecting $s[1]$ and $s[2]$ to $s[0]$:
    \begin{equation}
        s_l = \frac{||s[0]-s[1]|| + ||s[0]-s[2]||}{2}.
    \end{equation}
    \item \textbf{Angle offset ($s_o$):} The angular distance between the user's heading and the slice bisector:
    \begin{equation}
        s_o = |\theta - s_\beta|.
    \end{equation}
    \item \textbf{Slice area ($s_A$):} The area of the triangular slice $s$.
\end{itemize}
The composition of a slice is shown in \autoref{fig:slice_anatomy}.
The polygon $P$, and all of its slices, are returned at the end of \texttt{ComputeVisibilityPolygon($E, p, \theta$)}.


\subsubsection{\texorpdfstring{\texttt{GetActiveVirtualSlice($P_{virt},\theta_{virt}$)}}{}}
We define the ``active slice'' as the slice that the user is walking towards in the VE.
For this work, we assume that the user walks in the direction $\theta_{virt}$ that they are facing.
Thus, the active slice $s^{virt}$ is defined as:
\begin{equation}
    s^{virt} = \argmin_{s\in P_{virt}}|s_\beta - \theta_{virt}|.
\end{equation}
We return $s^{virt}$ at the end of \texttt{GetActiveVirtualSlice($P_{virt},\theta_{virt}$)}.

\subsubsection{\texorpdfstring{\texttt{GetMostSimilarPhysicalSlice($P_{phys},s^{virt}$)}}{}}
\label{subsubsec:get_most_similar_slice}
This function computes the slice $s^{phys} \in P_{phys}$ that is most similar to $s^{virt}$.
Here, we leverage our similarity metric to measure the alignment of slices in $P_{phys}$ to the active virtual slice $s^{virt}$.
Our slices are triangles, so we wish to compute a similarity metric that accurately measures how similar two triangles are.
Furthermore, the slices represent regions of free space that the user can walk in, so an ideal similarity metric would be able to compare two slices according to their similarity with regards to both shape \textit{and} navigability.
Measuring shape similarity is a very well-studied problem in geometric computing \cite{veltkamp2000shape}.
Likewise, the relationship between an environment's structure and navigation in that environment is well-studied \cite{conroy2001spatial,montello2005navigation,hillier1976space}.
Our goal is to design a metric that can compute the similarity of shapes with respect to the geometric structure as well as navigability. 
Therefore, we use a geometric measure of similarity (slice area), but we constrain the set of physical slices that we consider according to perceptual heuristics that tend to guide human locomotion (we only consider slices in the user's field of view).
We chose slice area over other shape similarity measures since we are primarily concerned with the user's proximity to $Obs_{phys}$, which is described by the slice's total area.

Given the physical visibility polygon $P_{phys}$, we first compute the set of eligible slices, which we will compare against $s^{virt}$.
This set, denoted $S^\dagger$, is defined as all physical slices for which the slice's bisector is less than $90^\circ$ away from the user's physical heading (this value is computed when $P_{phys}$ is constructed):
\begin{equation}
    S^\dagger = \{ s \in P_{phys} \ | \ s_{o} < \frac{\pi}{2} \}.
\end{equation}
Once $S^\dagger$ is computed, the physical slice that matches $s^{virt}$ most closely is simply the slice with the area closest to the area of $s^{virt}$:
\begin{equation}
    s^{phys} = \argmin_{ s \in S^\dagger}|s_A - s^{virt}_A|.
\end{equation}
This slice $s^{phys}$ is returned at the end of the function.

\subsubsection{\texorpdfstring{\texttt{GetRedirectionGains($s^{virt}, s^{phys}, \theta_{phys}$)}}{}}
\label{subsubsec:get_redir_gains}
We have now computed the region $s^{phys} \in E_{phys}$ that is most similar to the region $s^{virt} \in E_{virt}$ that the user is heading towards in virtual reality.
The final step is to set the rotation $g_r$, curvature $g_c$, and translation $g_t$ gains to steer the user towards $s^{phys}$.

An optimal direction vector, $v_o$, is defined as:
\begin{equation}
    v_o = \text{\texttt{unit\_vector(}}s^{phys}_\beta\texttt{)}.
\end{equation}
The \texttt{unit\_vector($\theta$)} function returns the vector $[\cos\theta, \sin\theta]^T$.
If the user is rotating in place, we apply a rotation gain $g_r$.
We set $g_r$ according to the following rule:
\begin{equation}
    g_r = 
    \begin{cases}
        minRotationGain & \text{user is turning away from }v_o, \\
        maxRotationGain & \text{user is turning towards }v_o, \\
    \end{cases}
\end{equation}
where $minRotationGain = 0.67$ and $maxRotationGain= 1.24$ \cite{steinicke2009estimation}.

If the user is walking, we apply translation and curvature gains to steer them in the direction of $v_o$.
Specifically, we set the curvature gain $g_c$ as:
\begin{equation}
    \begin{aligned}
        \theta_\Delta &= \texttt{signed\_angle(}\texttt{unit\_vector(}\theta_{phys}\texttt{)}, v_o\texttt{)} \\
        g_c &= \texttt{sign(}\theta_\Delta\texttt{)} \times maxCurvatureRadius.
    \end{aligned}
\end{equation}
Here, \texttt{signed\_angle($v_1, v_2$)} returns the positive angle between vectors $v_1$ and $v_2$ if the direction from $v_1$ to $v_2$ is counterclockwise.
Otherwise, it returns the negative angle between $v_1$ and $v_2$.
The \texttt{sign($\theta$)} function returns 1 or $-1$ depending on if $\theta$ is positive or negative.
We set $maxCurvatureRadius$ to $7.5m$, which is a commonly-used curvature threshold value in the RDW literature \cite{thomas2019general,azmandian2015physical,hodgson2013comparing}.

We set the translation gain $g_t$ according to the ratio between the average lengths of the physical and virtual slices, and we bound it by the perceptual thresholds for translation gains:
\begin{equation}
    g_t = \texttt{clamp(} s^{phys}_l / s^{virt}_l, minTransGain, maxTransGain \texttt{)}.
\end{equation}
Here, the \texttt{clamp(}$x,y,z$\texttt{)} function returns $x$, but ensures that it is greater than or equal to the lower bound $y$ and is less than or equal to the upper bound $z$.
We set $minTransGain = 0.86$ and $maxTransGain = 1.26$ since these are commonly-accepted translation gain thresholds \cite{thomas2019general,steinicke2009estimation}.

\label{sec:evaluation}
\section{Evaluation}
We conducted four experiments to evaluate the performance of our algorithm.
For each experiment, we compared our visibility-based algorithm against the alignment-based redirection controller (ARC) presented by Williams et al. \cite{williams2021arc}, the artificial potential field (APF) controller by Thomas et al. \cite{thomas2019general}, and the implementation of steer-to-center (S2C) by Hodgson et al. \cite{hodgson2013comparing}.
Our algorithm's reset function is the same as the one used by ARC \cite{williams2021arc}.
Both APF and S2C use the modified reset-to-center algorithm introduced by Thomas et al. \cite{thomas2019general}.
ARC is currently the best-performing reactive controller, while controllers based on potential-fields also perform fairly well.
We compare against S2C since it is a very common benchmark to compare against in the RDW literature; however, we note that S2C is not expected to do well in any of our experiments due to the obstacles present in the PE.
The first three experiments involve only static scenes with no dynamic obstacles.
Our fourth experiment investigates controller performance in dynamic scenes.

It is known that the structure of the environment affects a redirection controller's performance \cite{messinger2019effects}, so it is important to consider the environments' layouts when assessing a controller's efficacy.
There is currently no standard suite of test environments with which we can evaluate our algorithm, so we opted to test in environments used before to move towards a more standard set of environments.
Furthermore, we wish to test our algorithm in cluttered environments with many obstacles and narrow passageways, which are challenging scenarios where avoiding obstacles is non-trivial.
Thus, we evaluated our algorithm using two of the environment pairs from Williams et al. \cite{williams2021arc}.
When choosing our physical environments, we wished to test environments that did not represent traditional physical tracked spaces which often have very few obstacles.
Our goal is to test the viability of redirection in unstructured, irregular environments that may be commonly encountered in the real world, such as an office building with cubicles or a living room with tables and couches.

The performance metric we use is the number of resets incurred over the entire duration of the walked path since it is a fairly standard performance metric in the RDW community.
Another common metric is the average virtual distance walked between resets, but this metric is dependent on the number of resets, so it is slightly redundant to include both metrics in our evaluation.
Before conducting our experiments, we developed three hypotheses:
\begin{itemize}
    \item[\textbf{H1}]\label{H1} Our visibility-based steering algorithm will result in fewer collisions than the current state-of-the-art controllers in static scenes.
    \item[\textbf{H2}]\label{H2} Our visibility-based steering algorithm will result in fewer collisions than the current state-of-the-art controllers in dynamic scenes.
    \item[\textbf{H3}]\label{H3} Redirection controllers that use alignment will perform better in physical-virtual environment pairs that have more local similarity than they will in environment pairs that have less local similarity.
\end{itemize}

\subsection{Environment Pairs}
\begin{figure*}[t]
    \centering
    \includegraphics[width=.67\textwidth]{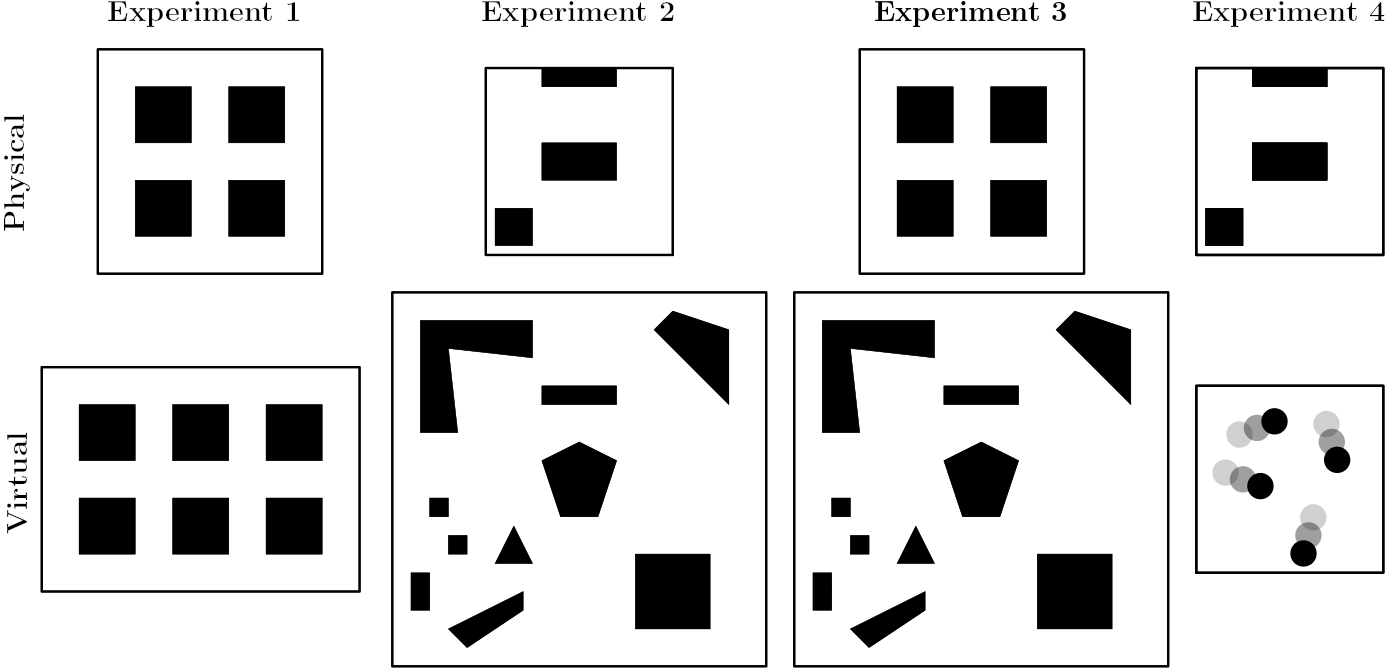}
    \caption{The layouts of the different environment pairs we tested in our experiments. The faded circles in the virtual environment for Experiment 4 indicate that the circles change position over time.}
    \label{fig:envs}
\end{figure*}

Each experiment had a different pair of physical and virtual environments.
The environments are shown in \autoref{fig:envs}.
Experiment 1 includes a $12m \times 12m$ physical and $17m \times 12m$ VE consisting of narrow corridors.
Experiment 2 has a $10m \times 10m$ PE with three rectangular obstacles and a $20m \times 20m$ VE with many convex and non-convex obstacles.
Using these two environments makes it easier to draw comparisons between our paper and that of Williams et al. \cite{williams2021arc}.
Experiment 3 uses the PE from Experiment 1 and the VE from Experiment 2.
We noticed that the pairs of environments used in Environment B and C in \cite{williams2021arc} (Experiments 1 and 2 here) appear to have a high degree of local similarity.
That is, the user's proximity to obstacles will be roughly similar between the physical and virtual environments.
In Environment B, both environments feature only narrow corridors and angular turns.
In Environment C, both physical and virtual environments have irregularity in the sizes and shapes of obstacles in both environments.
Therefore, for our third experiment, we opted to evaluate the controllers on a mixture between the environments from Experiments 1 and 2 since this would lower the degree of local similarity between the physical and virtual environments and allow us to test H3.

Finally, in Experiment 4, we used the physical environment from Experiment 2 and an empty $10m \times 10m$ environment with four dynamic, circular obstacles as the virtual environment.
We chose to use a VE with no static obstacles so that we could more easily study the influence of dynamic obstacles on the controller's performance.
That is, in an environment with both static and dynamic obstacles, it may be difficult to determine to what degree either type of obstacle influences the controller's performance.


\subsection{Simulated Environment}
To evaluate the effectiveness of our algorithm, we conducted extensive experiments with a simulated user walking in virtual reality.
Simulation has become a popular method of evaluation for redirection controllers since it allows researchers to quickly iterate on their algorithms and run large-scale experiments in a variety of environments \cite{strauss2020steering,williams2021arc,messinger2019effects,bachmann2019multi,chang2019redirection,dong2020dynamic,lee2019real,thomas2020towards,thomas2019general,thomas2020reactive}.
Our simulated user is represented as a circle with radius $0.5m$, and a reset is incurred whenever they came within $0.2m$ of any obstacle in the PE.
The user walked with a speed of $1m/s$ and turned with a speed of $90^\circ/s$.
Our simulation timestep size was $0.05$.

In our simulation, the model for generating the paths of the user is slightly different for static and dynamic scenes.
For static scenes, we used the motion model that was introduced by Azmandian et al. \cite{azmandian2015physical} and has been used by others \cite{thomas2020towards,williams2021arc}.
For dynamic scenes, we used ORCA \cite{van2011reciprocal} to generate trajectories for the user and the dynamic obstacles since ORCA generates smooth, collision-free paths for multiple objects in the same environment.

\subsection{Experiment Design}
For each of the static environment experiments, we generated 100 paths using the path model developed by Azmandian et al. \cite{azmandian2015physical}.
The average path length in these experiments was roughly $350m$.
We ran our simulation on these 100 paths once with each of the redirection controllers we evaluated (our visibility-based controller, ARC \cite{williams2021arc}, APF \cite{thomas2019general}, and S2C \cite{hodgson2013comparing}).
For each path, the user starts in a random location in the physical and virtual environments.
The user also has a random heading in both environments.
It is important to clarify that these random starting configurations were different between the 100 paths but were the same each time we simulated a particular path.

For the dynamic scene, we generated 100 collision-free paths for the user and the four dynamic obstacles in the VE using the ORCA \cite{van2011reciprocal}.
These paths had an average length of $136m$.
These paths were shorter than those used in static environments because we generated the dynamic paths to take roughly the same amount of timesteps to complete as in the static experiments, \textit{before} considering time taken for resets.
As we did in the static scenes, all redirection controllers were evaluated on the same 100 paths that we generated with ORCA.

\label{sec:results}
\section{Results}
We compared the number of resets across all 100 paths for the four algorithms that we tested in our experiments.
Some of our data violated assumptions of homoscedasticity or normality.
To account for these violated assumptions, we compared the controllers' performance with a robust one-way repeated measures ANOVA with $20\%$ trimmed means.
We used the WRS2 package in R to conduct our analyses \cite{mair2019robust}.
For all of the results presented in this section, we include the test statistic ($F$) and the significance level ($p$-value).
We also include the results of the post-hoc comparisons in \autoref{tab:results}, which are computed using linear contrasts.
For the post-hoc tests, we report the difference between the means ($\hat{\psi}$), the upper and lower confidence intervals ([CI~lower,~CI~upper]), and the significance level ($p$-value).
Note that we use confidence intervals as a measure of effect size \cite{lin2013research}.
The main focus of this work is to study the efficacy of our visibility-based algorithm, so we do not include the results of the post-hoc tests for any comparisons that do not include our visibility-based steering controller in the main text (e.g. ARC compared to APF).
Results comparing the other controllers can be found in the supplementary materials.

\setlength{\tabcolsep}{2.1pt}
\begin{table*}[ht]
    \centering
    \begin{tabular}{l|ccc|ccc|ccc|ccc}
     & \multicolumn{3}{c}{\textbf{Experiment 1}} & \multicolumn{3}{c}{\textbf{Experiment 2}} & \multicolumn{3}{c}{\textbf{Experiment 3}} & \multicolumn{3}{c}{\textbf{Experiment 4}}   \\ \hline
        Redirection Controller & $\hat{\psi}$ & CI & $p$ & $\hat{\psi}$ & CI & $p$ & $\hat{\psi}$ & CI & $p$ & $\hat{\psi}$ & CI & $p$  \\ \hline \hline
        Vis. Poly. vs ARC \cite{williams2021arc} & -32.7 & [28.9, 36.5] & $<.0001$ & -5.68 & [-2.73, -8.64] & $<.0001$ & -24.8 & [-20.6, -29.0] & $<.0001$ & -5.68 & [-3.70, -7.67] & $<.0001$  \\ \hline
        Vis. Poly. vs APF \cite{thomas2019general} & -76.4 & [-101, -51.7] & $<.0001$ & -105 & [-117, -92.1] & $<.0001$ & -49.6 & [-71.1, -28.0] & $<.0001$ & -64.3 & [-82.7, -45.9] & $<.0001$  \\ \hline
        Vis. Poly. vs S2C \cite{hodgson2013comparing} & -114 & [-143, -85.1] & $<.0001$ & -342 & [-362, -321] & $<.0001$ & -56.9 & [-71.7, -42.1] & $<.0001$ & -183 & [-203, -162] & $<.0001$  \\
    \end{tabular}
    \caption{The results of post-hoc pairwise comparisons between our visibility-based algorithm (denoted Vis. Poly.) and the other controllers (ARC, APF, S2C). The post-hoc tests are computed using linear contrasts. The $\hat{\psi}$ value is the average difference in means between our algorithm and the algorithm we compare against (ARC/APF/S2C). A negative $\hat{\psi}$ value indicates that our algorithm has a lower average number of resets across all 100 paths. The CI column presents the lower and upper bounds of the confidence interval, while the $p$ column presents the significance level of the difference between the algorithms. The $\hat{\psi}$ and CI values are rounded to three significant figures.}
    \label{tab:results}
\end{table*}






\subsection{Experiment 1}
\begin{figure}[t]
    \centering
    \includegraphics[width=.32\textwidth]{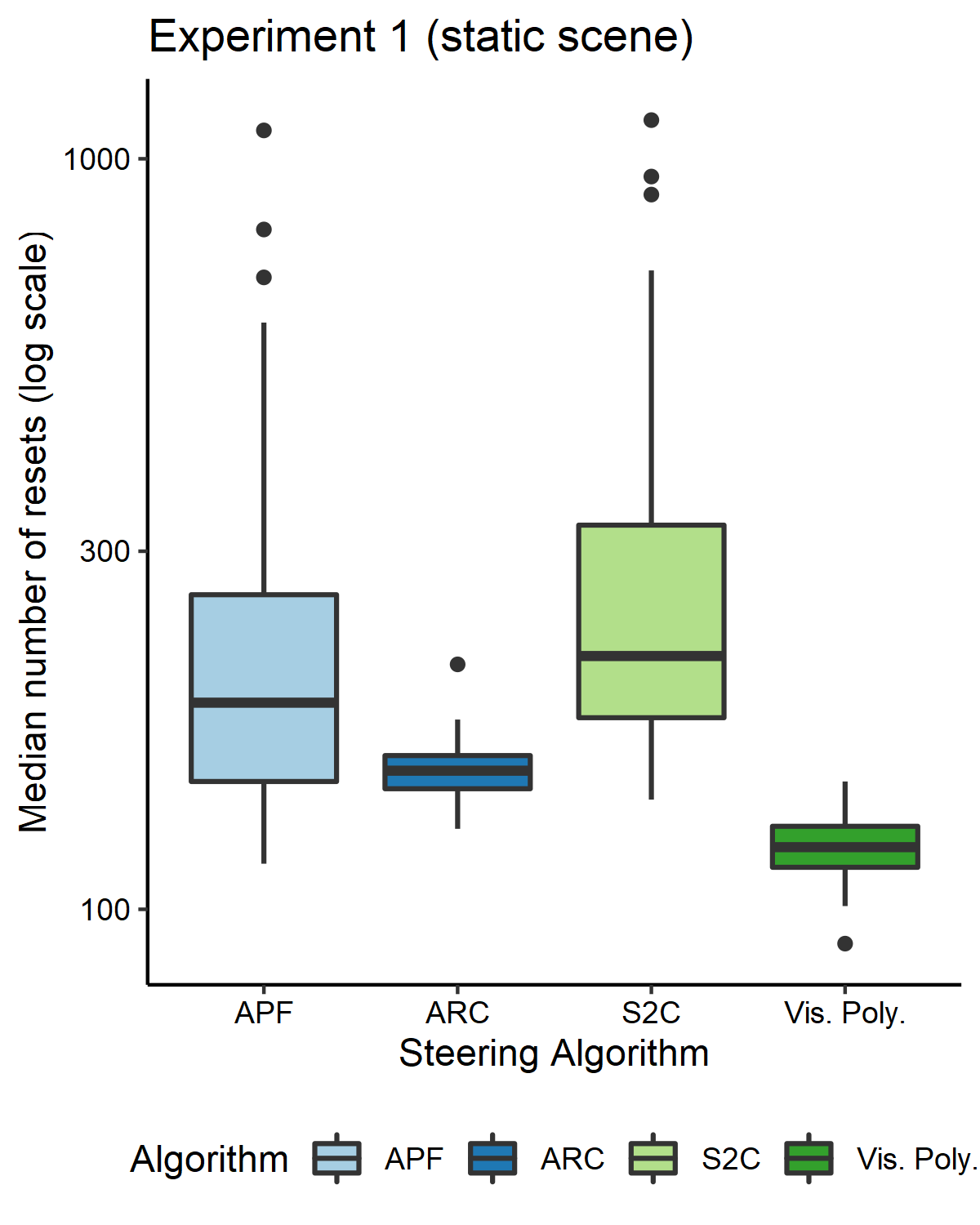}
    \caption{Boxplot of the number of resets for each algorithm, across all 100 paths in Experiment 1. Our visibility-based algorithm significantly outperformed each of the other redirection controllers.}
    \label{fig:exp1_boxplot}
\end{figure}

The robust ANOVA indicated a significant difference in the number of resets incurred when exploring with the different redirection controllers $F(1.88,111.03)=46.92, \ p<.0001$.
We see in \autoref{fig:exp1_boxplot} that our visibility-based algorithm achieves a median of 121 resets, which is 32 lower than the median of 153 achieved by ARC.
The median number of resets achieved by APF and S2C is slightly higher (188.5 and 217.5), and these two controllers have a much larger variance in the number of resets across all 100 paths.

\subsection{Experiment 2}
\begin{figure}[t]
    \centering
    \includegraphics[width=.32\textwidth]{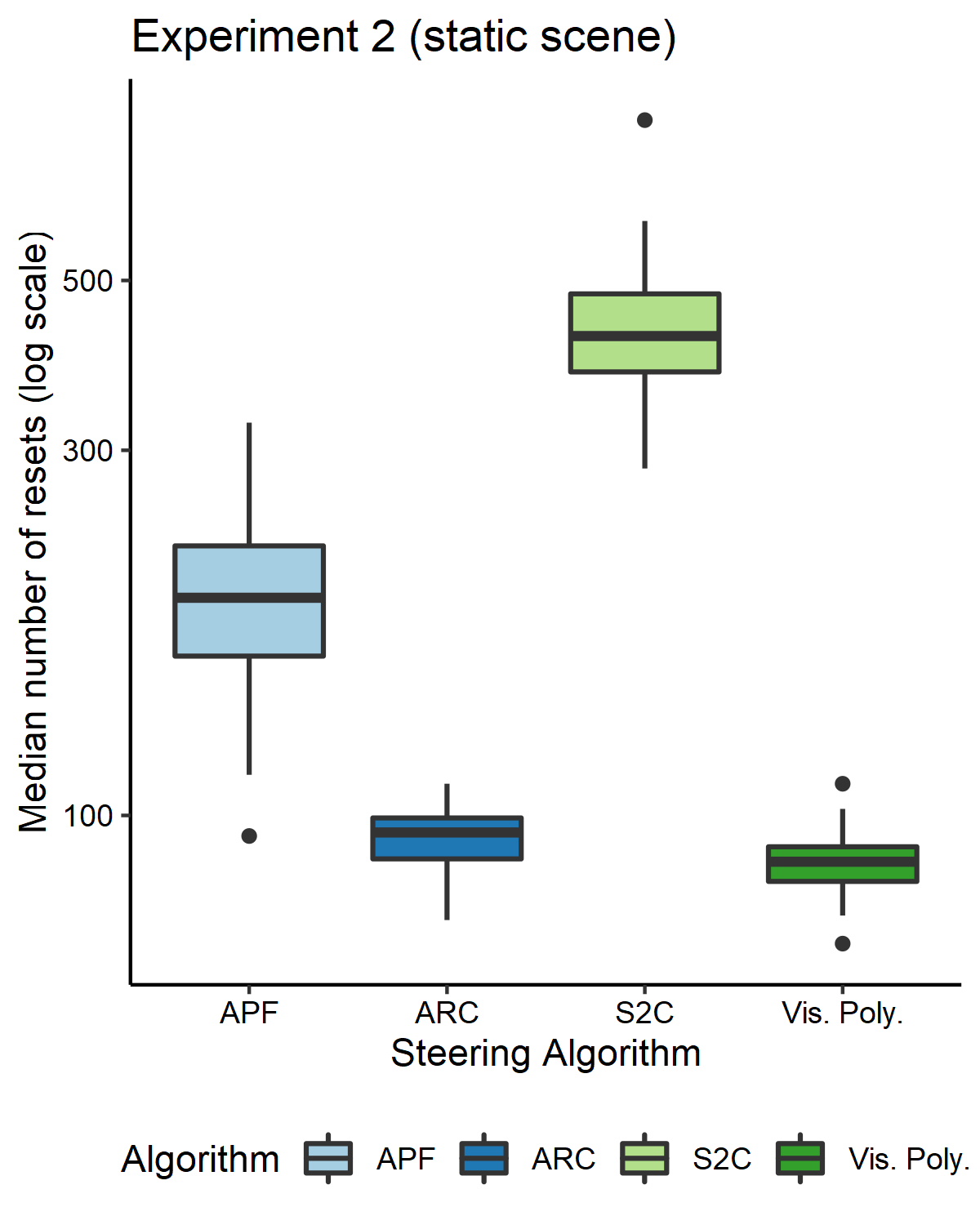}
    \caption{Boxplot of the number of resets for each algorithm, across all 100 paths in Experiment 2. The difference in the number of resets incurred is much larger for APF and S2C, which do not take advantage of alignment. ARC and our visibility-based controller (Vis. Poly.) have more similar performance levels, but our algorithm still produced significantly fewer resets.}
    \label{fig:exp2_boxplot}
\end{figure}

Our results showed a significant difference in performance between the RDW controllers ${F(1.66,97.65)=1266.94, \ p<.0001}$.
\autoref{fig:exp2_boxplot} shows us that the median number of resets incurred by our algorithm is 87, that the median for ARC is 95, and that APF and S2C achieve upwards of 150 median resets.

\subsection{Experiment 3}
\begin{figure}[t]
    \centering
    \includegraphics[width=.32\textwidth]{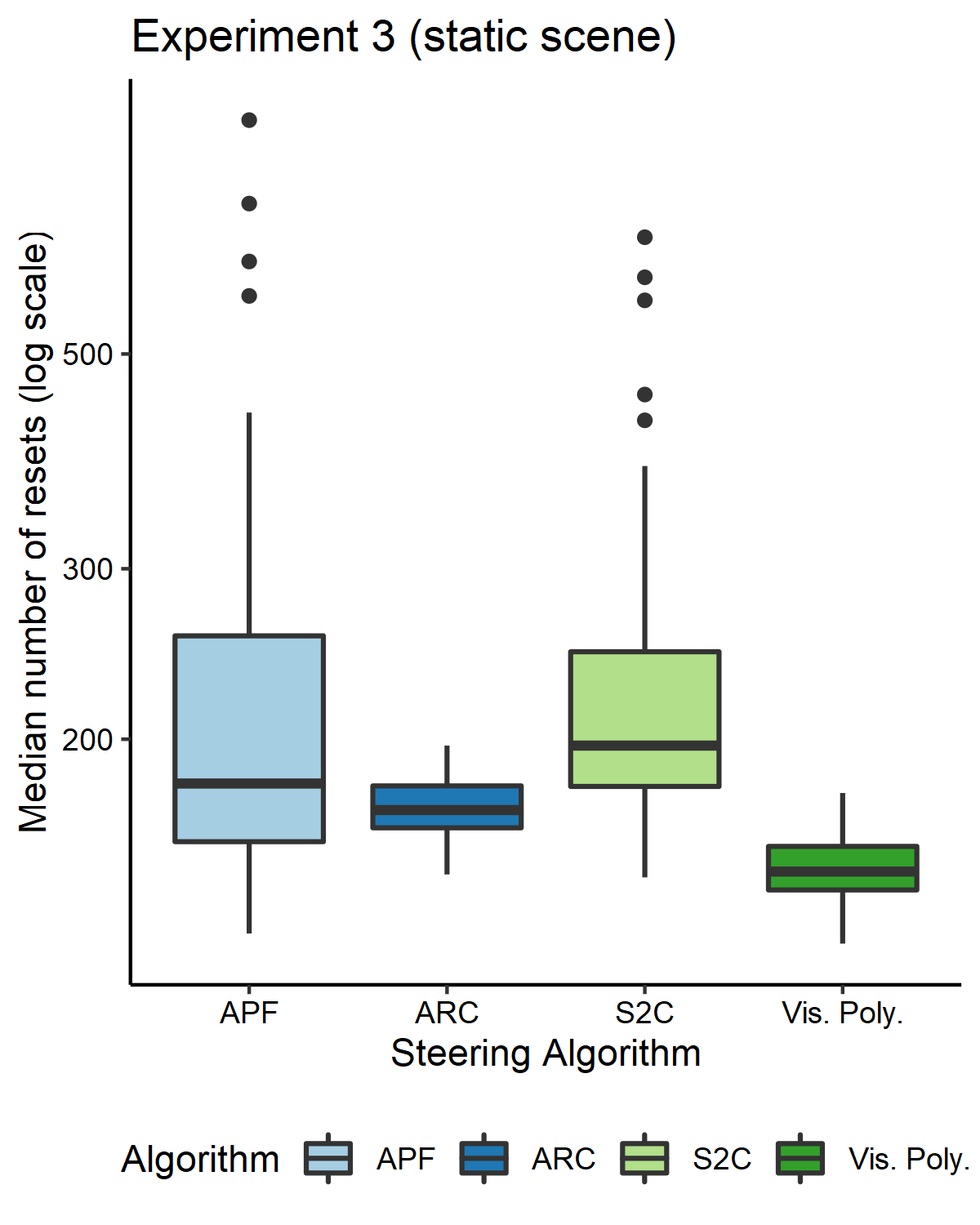}
    \caption{Boxplot of the number of resets for each algorithm, across all 100 paths in Experiment 3. Our controller based on visibility polygons performed significantly better than all other controllers.}
    \label{fig:exp3_boxplot}
\end{figure}

There was a significant difference between the four steering algorithms in terms of the number of resets incurred ${F(1.84,108.75)=29.90, \ p<.0001}$.
In \autoref{fig:exp3_boxplot} we see that this significant difference favors our new algorithm.
Similar to the pattern in first experiment, our visibility-based redirection controller achieves a median of 147.15 resets, which is about 25 fewer than that the 173 achieved by ARC, and APF and S2C have higher median resets (180 and 197) with a larger variance, too.

\subsection{Experiment 4}
In the dynamic scene, we found that the number of resets incurred was significantly different between the four controllers ${F(2.04,120.24)=319.03, \ p<.0001}$.
Once again, the significant difference favors the new visibility-based algorithm we presented in this paper (see \autoref{tab:results}).
Our algorithm had a median of 31 resets, while ARC had a median of 37 resets, APF had a median of 90 resets, and S2C had a median of 213.5 resets.

\begin{figure}[t]
    \centering
    \includegraphics[width=.32\textwidth]{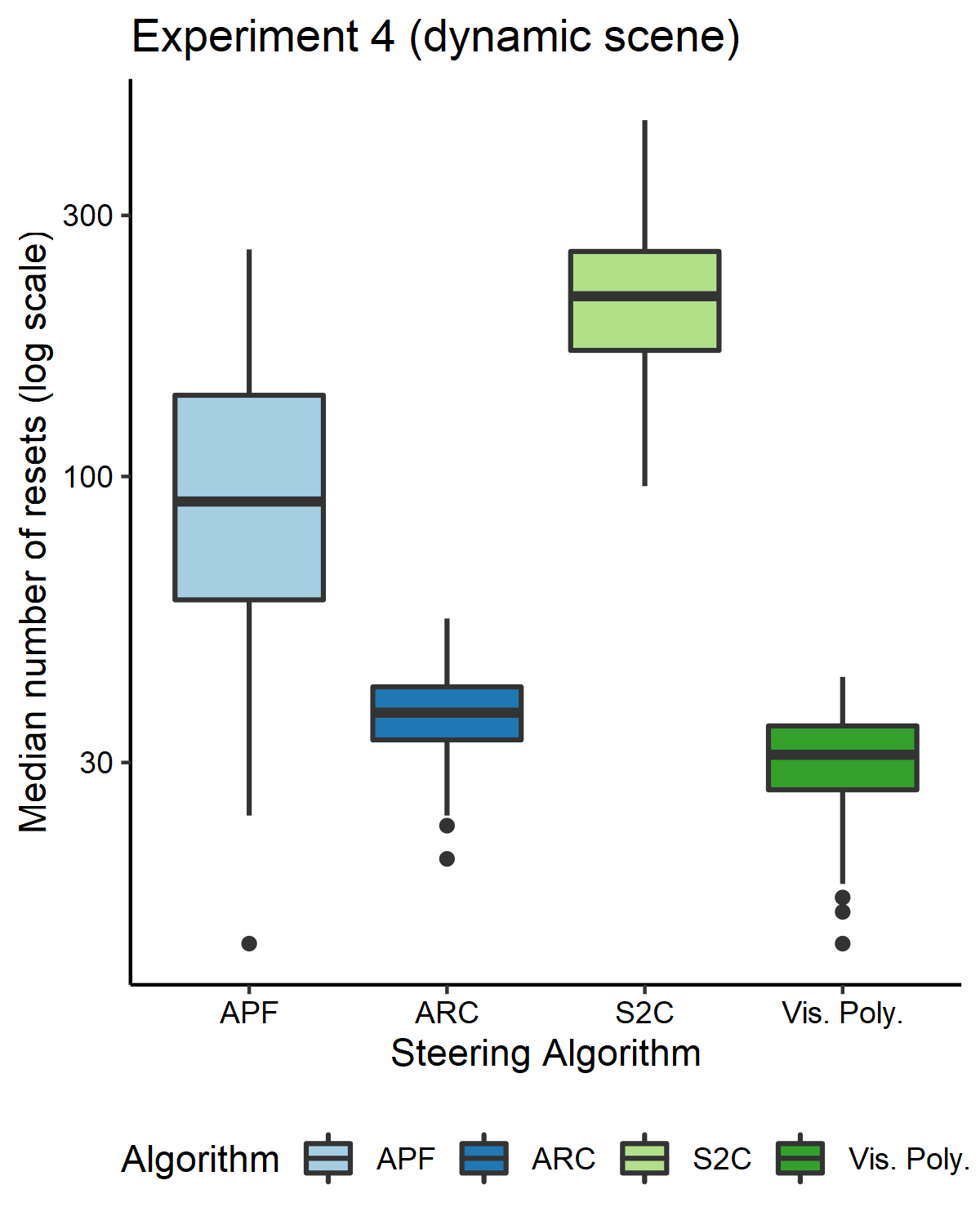}
    \caption{Boxplot of the number of resets for each algorithm, across all 100 paths in Experiment 4. In the dynamic scene we tested, we once again found that our visibility-based algorithm was significantly better than the other controllers at avoiding resets with physical obstacles.}
    \label{fig:exp4_boxplot}
\end{figure}

\section{Discussion}
\label{sec:Discussion}

\subsection{Static Scenes}
Our first three experiments were conducted in static scenes.
The results showed that our novel algorithm based on visibility polygons resulted in significantly fewer resets with obstacles in the environments we tested.
Overall, the results in the static scenes supported our first hypothesis (H1) that our visibility-based controller will outperform existing redirection controllers in static scenes.
In the first experiment, the physical and virtual environments had a considerable similarity since both environments featured regularly spaced narrow corridors, though the VE was larger than the physical one.
The simulated user incurred more resets in this PE than in the one used for Experiment 2.
This is likely because the narrow corridors make it difficult for the user to consistently walk without resets.

Experiment 3 used the PE from Experiment 1 and the virtual environment from Experiment 2.
We chose this combination in order to study the steering algorithms' performance in environment pairs that have little similarity on a local scale.
The PE is very regular and has only $90^\circ$ turns.
However, the VE contains many differently-shaped obstacles, creating irregular turns and regions of differing amounts of free space.
We found that all redirection controllers except APF performed worse in this experiment than they did in the others, which we believe is due to the more significant mismatch between the physical and virtual environments.
It suggests that controllers based on alignment will experience more difficulty avoiding resets in environment pairs with low local similarity than they will in pairs with high local similarity.
Note that APF, which does not steer the user with alignment, was able to perform \textit{better} in Experiment 3 than in Experiment 1.
The result of Experiment 3 supports our third hypothesis (H3) that controllers that steer based on alignment will perform better in environment pairs with high local similarity than in environment pairs with low local similarity.


\subsection{Dynamic Scenes}
In addition to testing our controller in static scenes, we also evaluated its performance in a scene with four dynamic virtual obstacles.
We found that our steering algorithm based on visibility polygons performed significantly better than ARC, APF, and S2C.
This result supports our second hypothesis (H2) that the additional information captured by the visibility polygon will lead to fewer resets in dynamic scenes.
One interesting result from the fourth experiment is that overall, the number of resets was much lower than in any of the other experiments.
This is initially surprising since dynamic environments intuitively seem to be more challenging to navigate.
However, considering that the paths for this experiment were generated using ORCA \cite{van2011reciprocal}, this is less surprising.
The paths in Experiments 1-3 consist only of straight-line segments, while the paths in Experiment 4 include curved path segments.
It may be the case that the curved segments are more amenable to redirection, which may be the cause of the lower number of resets seen in Experiment 4.
Another possible explanation for the lower number of resets is the fact that the user's speed varied in Experiment 4 but was constant in Experiments 1-3.
Finally, the paths in Experiment 4 were shorter than those in Experiments 1-3 since they were designed to take the same amount of time as the paths in the static scenes, and the user sometimes stopped walking in order to avoid collisions with the moving obstacles.


\subsection{Other Considerations}
In our algorithm, we used triangle area as our metric for measuring the similarity between slices of the physical and virtual visibility polygons (\autoref{subsubsec:get_most_similar_slice}).
Area is a relatively crude similarity metric, since two polygons can have very different shapes but still have the same area.
Metrics that can more accurately capture how similar two triangles are may lead to improvements in avoiding resets, since the improved metric will guide users towards free space regions that are more similar.
Shape similarity is a well-studied problem in the geometry and vision communities, so we believe that it is very likely that there exists a better triangle similarity metric.
There is likely room for more sophisticated similarity metrics that go beyond triangular slices.
We compared visibility polygons using the triangular slices that divide the polygons since it was a very natural way to segment the polygons, but it is possible that there exist better decompositions for comparing visibility polygons.

\vspace{0em}
\label{sec:conclusion}
\section{Conclusion, Limitations, and Future Work}

In this paper, we presented a novel formulation of the redirected walking problem and developed a visibility-based redirection controller that takes advantage of this formulation to yield fewer resets during walking.
Our formal description of the redirection problem frames it in terms of optimizing the redirection gains such that the user follows a collision-free path in the physical free space while simultaneously traveling on a collision-free path in the virtual free space.
We hope that this new formulation will allow researchers to further improve redirection controllers by leveraging motion planning techniques that have been well-studied by the robotics community.

The novel redirection controller that we also presented in this work is based on the key insight that the visibility polygon provides a reliable representation of the user's free space at their location.
By dividing the physical and virtual visibility polygons into ``slices'' and steering the user towards a physical slice that is similar to the current virtual slice, we were able to avoid significantly more resets than the state-of-the-art controllers were able to.
As evidenced by our simulation-based results, our algorithm proved to be more effective in both static and dynamic scenes where the physical and virtual environments had locally dissimilar layouts.

Although our results were positive, it is important to discuss the limitations of this work, too.
In our experiments, we only evaluated the algorithms in simulated environments.
Simulation can be effective for quickly getting an understanding of a RDW algorithm's strengths and weaknesses, but full user studies should be conducted to gain a more complete evaluation of a controller's efficacy and to highlight shortcomings that may not arise in simulations.
Another limitation of this work is that we only tested scenarios involving one user in the PE and VE.
Many use-cases for VR involve multiple users, either in the same PE or different PEs.
Thus, it is important to continue developing RDW controllers that can improve the locomotion experience for multiple users \cite{azmandian2017evaluation, bachmann2019multi, dong2020dynamic}.
Finally, it is important to consider the challenges involved in extending our algorithm to commodity VR hardware and uncontrolled, irregular physical and virtual environments.
Our algorithm requires knowledge of the locations of obstacles around the user in the PE and VE.
Obtaining this data for the VE is generally not problematic, since the virtual environment data is already being used for rendering purposes.
However, collecting layout data of the PE is considerably more difficult, since this involves object detection and tracking in real time.
This problem can be sidestepped by providing the RDW controller with a map of the PE if it is known beforehand, but this cannot be guaranteed in a commodity VR setting where the application will be used in a variety of PEs.
In addition to collecting environment data, it is important to consider the run-time of the algorithm.
Visibility computations are very well-studied problems in computer graphics and robotics, so a considerable amount of effort has been invested into developing efficient and robust algorithms for computing visibility.
Nevertheless, since many other computations need to be done on each timestep in VR applications, computing visibility polygons in real-time may be difficult if the environment contains a large number of obstacles.

Future work should investigate different ways to use the visibility polygon for more sophisticated steering, such as more accurate measures of shape similarity.
Our results showed that for some environments, the performance improvement afforded by visibility polygons was not large (though it was statistically significant).
It may be the case that our algorithm does not use visibility polygons to their full potential.
There is not a lot of research studying the relationship between the shape of the virtual path and a controller's performance, so the impact of the path models on a controller's performance is an open question that warrants more research.
Another interesting area for future work is to extend our visibility-based controller to real-world scenarios where the exact geometry of the environments is not known.
Our experiments were conducted using reliable simulations to show that our algorithm is effective, but calculating visibility polygons in real PEs will likely introduce new challenges.
Finally, extensive in-person user studies should be conducted now that we know that our visibility-based controller can yield significant benefits.


\bibliographystyle{abbrv-doi}

\bibliography{template}

\newpage 
\newpage
\clearpage

\renewcommand{\thesection}{\Alph{section}}
\setcounter{section}{0}
\section{Supplementary Materials} \label{sec:supplementary}

\subsection{Environment Layouts}

\subsubsection{Experiment 1}
\begin{table}[hb]
    \centering
    \begin{tabular}{c|p{53mm}}
        \multicolumn{2}{c}{\textbf{Experiment 1 (physical)}}    \\ 
        Boundary & $(-6, -6), (6, -6), (6, 6), (-6, 6)$ \\ \hline
        Obstacle 1 & $(-4, -4), (-1, -4), (-1, -1), (-4, -1)$ \\ \hline
        Obstacle 2 & $(1, -4), (4, -4), (4, -1), (1, -1)$ \\ \hline
        Obstacle 3 & $(1, 1), (4, 1), (4, 4), (1, 4)$ \\ \hline
        Obstacle 4 & $(-4, 1), (-1, 1), (-1, 4), (-4, 4)$ \\ 
        \multicolumn{2}{c}{}    \\ 
        \multicolumn{2}{c}{\textbf{Experiment 1 (virtual)}}    \\ 
        Boundary & $(-11, -6), (6, -6), (6, 6), (-11, 6)$ \\ \hline
        Obstacle 1 & $(-4, -4), (-1, -4), (-1, -1), (-4, -1)$ \\ \hline
        Obstacle 2 & $(1, -4), (4, -4), (4, -1), (1, -1)$ \\ \hline
        Obstacle 3 & $(1, 1), (4, 1), (4, 4), (1, 4)$ \\ \hline
        Obstacle 4 & $(-4, 1), (-1, 1), (-1, 4), (-4, 4)$ \\ \hline
        Obstacle 5 & $(-9, 1), (-6, 1), (-6, 4), (-9, 4)$ \\ \hline
        Obstacle 6 & $(-9, -4), (-6, -4), (-6, -1), (-9, -1)$ \\ 
    \end{tabular}
    \label{tab:env_config1}
\end{table}

\subsubsection{Experiment 2}
\begin{table}[hb]
    \centering
    \begin{tabular}{c|p{53mm}}
        \multicolumn{2}{c}{\textbf{Experiment 2 (physical)}}    \\ 
        Boundary & $(-5, -5), (5, -5), (5, 5), (-5, 5)$ \\ \hline
        Obstacle 1 & $(-4.5, -4.5), (-2.5, -4.5), $ \newline $  (-2.5, -2.5), (-4.5, -2.5)$ \\ \hline
        Obstacle 2 & $(-2, -1), (2, -1), (2, 1), (-2, 1)$ \\ \hline
        Obstacle 3 & $(-2, 4), (2, 4), (2, 5), (-2, 5)$ \\ 
        \multicolumn{2}{c}{}    \\ 
        \multicolumn{2}{c}{\textbf{Experiment 2 (virtual)}}    \\ 
        Boundary & $(10, -10), (10, 10), (-10, 10), (-10, -10)$ \\ \hline
        Obstacle 1 & $(-4.5, -4.5), (-2.5, -4.5), (-3.5, -2.5)$ \\ \hline
        Obstacle 2 & $(0, 2), (2, 1), (1, -2), (-1, -2), (-2, 1)$ \\ \hline
        Obstacle 3 & $ (-2,4), (2,4), (2,5), (-2,5) $ \\ \hline
        Obstacle 4 & $ (-8.5,8.5), (-8.5,2.5), (-6.5,2.5), $ \newline $ (-7,7), (-2.5,6.5), (-2.5,8.5) $ \\ \hline
        Obstacle 5 & $ (-8,-1), (-8,-2), (-7,-2), (-7,-1) $ \\ \hline
        Obstacle 6 & $ (-7,-3), (-7,-4), (-6,-4), (-6,-3) $ \\ \hline
        Obstacle 7 & $ (-9,-5), (-9,-7), (-8,-7), (-8,-5) $ \\ \hline
        Obstacle 8 & $ (-6,-9), (-3,-7), (-3,-6), (-7,-8) $ \\ \hline
        Obstacle 9 & $ (3,-4), (3,-8), (7,-8), (7,-4) $ \\ \hline
        Obstacle 10 & $ (5,9), (4,8), (8,4), (8,8) $ \\
    \end{tabular}
    \label{tab:env_config2}
\end{table}

\newpage

\subsubsection{Experiment 3}
\begin{table}[hb]
    \centering
    \begin{tabular}{c|p{53mm}}
        \multicolumn{2}{c}{\textbf{Experiment 3 (physical)}}    \\ 
        Boundary & $(-6, -6), (6, -6), (6, 6), (-6, 6)$ \\ \hline
        Obstacle 1 & $(-4, -4), (-1, -4), (-1, -1), (-4, -1)$ \\ \hline
        Obstacle 2 & $(1, -4), (4, -4), (4, -1), (1, -1)$ \\ \hline
        Obstacle 3 & $(1, 1), (4, 1), (4, 4), (1, 4)$ \\ \hline
        Obstacle 4 & $(-4, 1), (-1, 1), (-1, 4), (-4, 4)$ \\ 
        \multicolumn{2}{c}{}    \\ 
        \multicolumn{2}{c}{\textbf{Experiment 3 (virtual)}}    \\ 
        Boundary & $(10, -10), (10, 10), (-10, 10), (-10, -10)$ \\ \hline
        Obstacle 1 & $(-4.5, -4.5), (-2.5, -4.5), (-3.5, -2.5)$ \\ \hline
        Obstacle 2 & $(0, 2), (2, 1), (1, -2), (-1, -2), (-2, 1)$ \\ \hline
        Obstacle 3 & $ (-2,4), (2,4), (2,5), (-2,5) $ \\ \hline
        Obstacle 4 & $ (-8.5,8.5), (-8.5,2.5), (-6.5,2.5), $ \newline $ (-7,7), (-2.5,6.5), (-2.5,8.5) $ \\ \hline
        Obstacle 5 & $ (-8,-1), (-8,-2), (-7,-2), (-7,-1) $ \\ \hline
        Obstacle 6 & $ (-7,-3), (-7,-4), (-6,-4), (-6,-3) $ \\ \hline
        Obstacle 7 & $ (-9,-5), (-9,-7), (-8,-7), (-8,-5) $ \\ \hline
        Obstacle 8 & $ (-6,-9), (-3,-7), (-3,-6), (-7,-8) $ \\ \hline
        Obstacle 9 & $ (3,-4), (3,-8), (7,-8), (7,-4) $ \\ \hline
        Obstacle 10 & $ (5,9), (4,8), (8,4), (8,8) $ \\
    \end{tabular}
    \label{tab:env_config3}
\end{table}

\subsubsection{Experiment 4}
\begin{table}[hb]
    \centering
    \begin{tabular}{c|p{53mm}}
        \multicolumn{2}{c}{\textbf{Experiment 4 (physical)}}    \\ 
        Boundary & $(-5, -5), (5, -5), (5, 5), (-5, 5)$ \\ \hline
        Obstacle 1 & $(-4.5, -4.5), (-2.5, -4.5), $ \newline $  (-2.5, -2.5), (-4.5, -2.5)$ \\ \hline
        Obstacle 2 & $(-2, -1), (2, -1), (2, 1), (-2, 1)$ \\ \hline
        Obstacle 3 & $(-2, 4), (2, 4), (2, 5), (-2, 5)$ \\ 
        \multicolumn{2}{c}{}    \\ 
        \multicolumn{2}{c}{\textbf{Experiment 4 (virtual)}}    \\ 
        Boundary & $(-11, -6), (6, -6), (6, 6), (-11, 6)$ \\ 
    \end{tabular}
    \label{tab:env_config}
\end{table}

\clearpage

\subsection{Additional Results}

\subsubsection{Post-hoc Comparisons For Number Of Resets}
Here we show the results of post-hoc comparisons for total number of resets between the other controllers we tested against in our experiments.

\setlength{\tabcolsep}{2.1pt}
\begin{table}[!htbp]
    \centering
    \begin{tabular}{l|ccc|ccc|ccc|ccc}
    \multicolumn{13}{c}{\textbf{Number of Resets}} \\ 
     & \multicolumn{3}{c}{\textbf{Experiment 1}} & \multicolumn{3}{c}{\textbf{Experiment 2}} & \multicolumn{3}{c}{\textbf{Experiment 3}} & \multicolumn{3}{c}{\textbf{Experiment 4}}   \\ \hline
        Redirection Controller & $\hat{\psi}$ & CI & $p$ & $\hat{\psi}$ & CI & $p$ & $\hat{\psi}$ & CI & $p$ & $\hat{\psi}$ & CI & $p$  \\ \hline \hline
        ARC \cite{williams2021arc} vs APF \cite{thomas2019general} & -42.3 & [-66.6, -18.1] & $<.0001$ & -99.0 & [-112, -86.4] & $<.0001$ & -25.2 & [-46.8, -3.56] & $<.01$ & -59.1 & [-77.4, -40.7] & $<.0001$ \\ \hline
        ARC \cite{williams2021arc} vs S2C \cite{hodgson2013comparing} & -82.4 & [-114, -51.3] & $<.0001$ & -335 & [-356, -313] & $<.0001$ & -32.0 & [-47.6, -16.3] & $<.0001$ & -177 & [-197, -157] & $<.0001$  \\ \hline
        APF \cite{thomas2019general} vs S2C \cite{hodgson2013comparing} & -35.2 & [-71.3, 0.882] & $<.01$ & -233 & [-254, -212] & $<.0001$ & -8.02 & [-31.3, 15.2] & $=.350$ & -112 & [-133, -91.5] & $<.0001$  \\
    \end{tabular}
    \label{tab:results2}
\end{table}

\subsubsection{Post-hoc Comparisons For Resets Per Meter}
The table below shows the results of post-hoc comparisons between controllers according to the number of resets incurred per meter walked in the virtual environment.
\begin{table}[ht]
    \setlength{\tabcolsep}{1pt}
    \centering
    \begin{tabular}{l|ccc|ccc|ccc|ccc}
    \multicolumn{13}{c}{\textbf{Resets per Meter}} \\ 
     & \multicolumn{3}{c}{\textbf{Experiment 1}} & \multicolumn{3}{c}{\textbf{Experiment 2}} & \multicolumn{3}{c}{\textbf{Experiment 3}} & \multicolumn{3}{c}{\textbf{Experiment 4}}   \\ \hline
        Redirection Controller & $\hat{\psi}$ & CI & $p$ & $\hat{\psi}$ & CI & $p$ & $\hat{\psi}$ & CI & $p$ & $\hat{\psi}$ & CI & $p$  \\ \hline \hline
        Vis. Poly. vs ARC \cite{williams2021arc} & -.0929 & [-.0811, -.105] & $<.0001$ & -.0157 & [-.00746, -.0239] & $<.0001$ & -.0688 & [-.0570, -.0806] & $<.0001$ & -.0428 & [-.0278, -.0577] & $<.0001$ \\ \hline
        Vis. Poly. vs APF \cite{thomas2019general} & -.220 & [-.291, -.149] & $<.0001$ & -.299 & [-.324, -.255] & $<.0001$ & -.137 & [-.198, -.0767] & $<.0001$ & -.471 & [-.598, -.345] & $<.0001$ \\ \hline
        Vis. Poly. vs S2C \cite{hodgson2013comparing} & -.327 & [-.411, -.242] & $<.0001$ & -.940 & [-.995, -.885] & $<.0001$ & -.158 & [-.200, -.116] & $<.0001$ & -1.34 & [-1.47, -1.21] & $<.0001$ \\ \hline
        ARC \cite{williams2021arc} vs APF \cite{thomas2019general} & -.123 & [-.195, -.0515] & $<.0001$ & -.274 & [-.309, -.239] & $<.0001$ & -.0697 & [.-131, -.00858] & $<.01$ & -.431 & [-.556, -.306] & $<.0001$ \\ \hline
        ARC \cite{williams2021arc} vs S2C \cite{hodgson2013comparing} & -.237 & [-.325, -.149] & $<.0001$ & -.920 & [-.975, -.864] & $<.0001$ & -.0886 & [-.132, -.0456] & $<.0001$ & -1.30 & [-1.43, -1.17] & $<.0001$ \\ \hline
        APF \cite{thomas2019general} vs S2C \cite{hodgson2013comparing} & -.101 & [-.207, .00495] & $<.05$ & -.642 & [-.700, -.587] & $<.0001$ & -.0220 & [-.0851, .0412] & $=.346$ & -.829 & [-.975, -.683] & $<.0001$ \\
    \end{tabular}
    \label{tab:results3}
\end{table}

\end{document}